\documentclass[letterpaper, 10 pt, conference]{ieeeconf}  % Comment this line out if you need a4paper

\IEEEoverridecommandlockouts                              % This command is only needed if 
\usepackage{graphics} % for pdf, bitmapped graphics files
\usepackage{epsfig} % for postscript graphics files
\usepackage{mathptmx} % assumes new font selection scheme installed
\usepackage{times} % assumes new font selection scheme installed
\usepackage{amsmath} % assumes amsmath package installed
\usepackage{amssymb}  % assumes amsmath package installed
\usepackage{xcolor}
\usepackage{bm}
\usepackage{caption}
\usepackage{float}
\usepackage[margin = 1in]{geometry}
\usepackage{media9}
\usepackage{subfig}
\usepackage{graphicx}
\usepackage{comment}
\usepackage{multicol}
\usepackage{float}
\usepackage{url}
\usepackage{booktabs}
\usepackage{makecell}

\newcommand{\vp}{\mathbf{p}}

\newcommand{\vv}{\mathbf{v}}

\newcommand{\va}{\mathbf{a}}

\newcommand{\vb}{\mathbf{b}}

\newcommand{\veta}{\bm{\eta}}

\newcommand{\vomega}{\bm{\omega}}
\newcommand{\vmu}{\bm{\mu}}
\newcommand{\vSigma}{\bm{\Sigma}}

\title{\LARGE \bf
	Quantifying the Sim2real Gap for GPS and IMU Sensors
}

\author{Ishaan Mahajan$^{1}$, Huzaifa Unjhawala$^{1}$, Harry Zhang$^{1}$, Zhenhao Zhou$^{1}$, Aaron Young$^{2}$, Alexis Ruiz$^{1}$,  \\ 
Stefan Caldararu$^{1}$,  Nevindu Batagoda$^{1}$, Sriram Ashokkumar$^{1}$, and Dan Negrut$^{1}$% <-this % stops a space
\thanks{$^{1}$University of Wisconsin-Madison, Madison, WI 53706, USA
        {\tt\small \{imahajan, unjhawala, hzhang699, zzhou292, aruiz26, scaldararu, batagoda, ashokkumar2, negrut\}@wisc.edu}}%
\thanks{$^{2}$Massachusetts Institute of Technology, Cambridge, MA 02139, USA
        {\tt\small aryoung@mit.edu}}%
}

\begin{document}

%  Notes command (uncomment the first for publication, uncomment the second
%  when working on the paper)

%\definecolor{darkgreen}{rgb}{0,0.4,0}
\newcommand\SBELcomment[1]{{\textcolor{red}{\bf{#1}}}}
\definecolor{arsenic}{rgb}{0.23, 0.27, 0.29}
\definecolor{charcoal}{rgb}{0.21, 0.27, 0.31}
\definecolor{hanblue}{rgb}{0.27, 0.42, 0.81}
\definecolor{blue-ncs}{rgb}{0.0, 0.53, 0.74}
\definecolor{awesome}{rgb}{1.0, 0.13,0.32}
\definecolor{darkgreen}{rgb}{0, .4,0}

\newcommand{\firstReviewer}[1]{{\textcolor{hanblue}{{#1}}}}
\newcommand{\secondReviewer}[1]{{\textcolor{darkgreen}{{#1}}}}
\newcommand{\thirdReviewer}[1]{{\textcolor{blue-ncs}{{#1}}}}
\newcommand{\updatedText}[1]{{\textcolor{red}{{#1}}}}

\newcommand{\CHRONO}{{\sffamily{{Chrono}}}}
\newcommand{\ChronoVehicle}{{\softpackage{{Chrono}}}::Vehicle}
\newcommand{\softpackage}[1]{{\sffamily{#1}}}

\newcommand{\cA}{{\mathcal A}}
\newcommand{\cL}{{\mathcal L}}
\newcommand{\cone}{{\Upsilon}}
\newcommand{\cD}{{\mathcal D}}

\newcommand{\norm}[1]{{ \left| { \left| #1 \right| }  \right| }}
\newcommand{\subVect}[2]{{#1}_{{#2}}}

\newcommand{\vect}[1]{\mathbf{#1}}
\newcommand{\matr}[1]{\mathbf{#1}}

\maketitle
\thispagestyle{empty}
\pagestyle{empty}

%%%%%%%%%%%%%%%%%%%%%%%%%%%%%%%%%%%%%%%%%%%%%%%%%%%%%%%%%%%%%%%%%%%%%%%%%%%%%%%%

\begin{abstract}
Simulation can and should play a critical role in the development and testing of algorithms for autonomous agents. What might reduce its impact is the ``sim2real'' gap -- the algorithm response differs between operation in simulated versus real-world environments. This paper introduces an approach to evaluate this gap, focusing on the accuracy of sensor simulation -- specifically IMU and GPS -- in velocity estimation tasks for autonomous agents. Using a scaled autonomous vehicle, we conduct 40 real-world experiments across diverse environments then replicate the experiments in simulation with five distinct sensor noise models. We note that direct comparison of raw simulation and real sensor data fails to quantify the sim2real gap for robotics applications. We demonstrate that by using a state of the art state-estimation package as a ``judge'', and by evaluating the performance of this state-estimator in both real and simulated scenarios, we can isolate the sim2real discrepancies stemming from sensor simulations alone. The dataset generated is open-source and publicly available for unfettered use.

\end{abstract}
\begin{figure}
        \centering
        \includegraphics[width=0.47\textwidth]{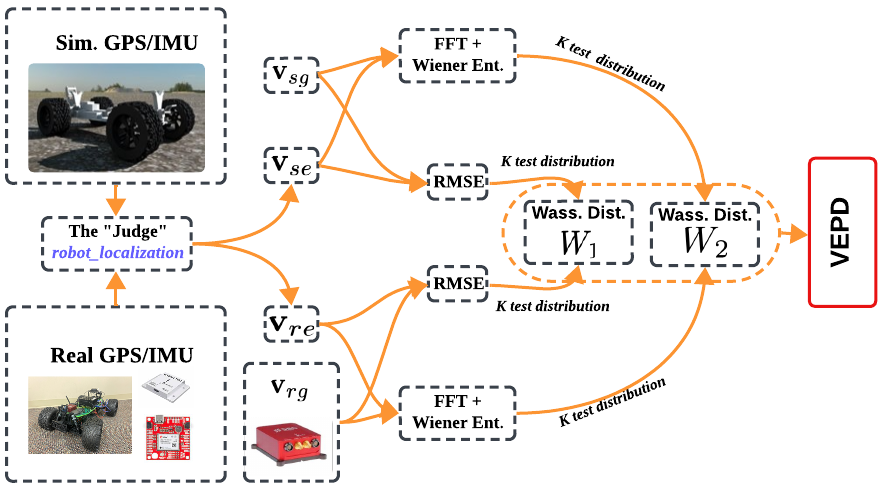}
        \caption{The state estimator serves as the ``judge,'' processing both real-world and simulated sensor data to produce velocity estimates $\vv_{re}$ and $\vv_{se}$, respectively. In the real world, ground-truth velocity ($\vv_{rg}$) is provided by a validated INS system, while the simulator yields $\vv_{sg}$.  RMSE values ($E_r$ and $E_s$) are calculated for both the real and simulated cases (Eq.~\ref{eq:RMSE}). Additionally, Fast Fourier Transforms (FFTs) are employed to compute Wiener Entropy values $S_r$ and $S_s$ (Eq.~\ref{eq:ent_diff}). This process is repeated across K tests, generating sets $A_1$, $A_2$, $B_1$, and $B_2$.  The Wasserstein distance $W_1(A_1,B_1)$ and $W_2(A_2,B_2)$ compares these sets, yielding the Velocity Estimation Performance Difference (VEPD) score as an average.}
        \label{fig:flow_chart}
        \vspace{-10pt}
\end{figure}

%%%%%%%%%%%%%%%%%%%%%%%%%%%%%%%%%%%%%%%%%%%%%%%%%%%%%%%%%%%%%%%%%%%%%%%%%%%%%%%%
\section{INTRODUCTION}
\label{sec:intro}

A prerequisite for the safe operation of many autonomous systems is robust and accurate localization within complex environments. Global Positioning Systems (GPS) with Inertial Measurement Units (IMU) are often relied upon for achieving this goal~\cite{Ding2020LongitudinalVS,Gu2020NonlinearOD}. This sensor fusion strategy compensates for the individual limitations of GPS and IMUs. Although GPS provides global positioning, its reliability and precision can be compromised in areas of signal obstruction. IMUs provide high-frequency motion data; however, they are prone to drift and magnetic interference, leading to accumulated errors over time~\cite{Sukkarieh1999HighIntegrityIMU}. Typically, Kalman filters or particle filters fuse GPS and IMU measurements while accounting for their individual uncertainties. However, developing and fine-tuning state estimators is a complex task; real-world testing demands significant engineering effort, which is both time-consuming and costly. Moreover, real-world validation is often constrained by the availability of accurate, noise-free ground truth data, making it difficult to validate and refine these systems effectively. Simulation environments alleviate this by providing a controlled space where various scenarios can be tested thoroughly. Through simulation, developers can iteratively test and refine their state estimation algorithms, improving their robustness and reliability before deployment in real-world applications~\cite{Saeedmanesh2021,Jaradat2017NonLinearAD}. However, the effectiveness of using simulation to validate state estimation algorithms is contingent on the accuracy of the simulated sensor data. Inaccurate sensor models can lead to a misrepresentation of the real-world sensor behavior~\cite{kadian2020sim2real}, resulting in unreliable state estimation algorithms that contribute to the sim2real gap. 

Comparing synthetic (simulated) vs. real raw sensor data, which is a standard evaluation technique for sensor model fidelity, does not fully capture the sim2real gap's impact on downstream tasks~\cite{Manivasagam2023LiDARSim2Real}~\cite{Wang2008GPSModelling}. Raw comparison neglects how sensor data is ultimately utilized.  To address this, our methodology assesses the gap's impact by zeroing in on the very task the sensor data is used in, i.e., velocity state estimation. We focus on five variants of sensor models, validated against 40 real-world tests.  

\subsection{Related Work}
Simulation can accelerate the development and testing of robotic systems~\cite{PNASsimRobotics2021,karenSimRobotics2021,Xuemin2023Sim2RealSurvey}, and a range of robotics simulators are available on the market~\cite{todorovComparisonSimEngines2015,KoenigDesign2004,mujoco2012,Pybullet2023,IssacNvidia2023,chronoOverview2016}. A recent simulation-in-robotics survey of 82 robotics developers found that 84\% of respondents use simulation for testing and validation of algorithms~\cite{Afzal2021SimulationRoboticsTestAutomation}. The survey also found that the most common challenge faced by developers is related to the \textit{sim2real} gap. A brief summary of the state of the art on the identification, quantification, and mitigation of the sim2real gap specifically for sensor simulation is provided below.

\noindent {\textbf{\textit{Sim2Real} Gap in Sensor Simulation}}. The \textit{sim2real} gap is particularly pronounced in sensor simulation, where the accuracy and fidelity of the synthetic data is crucial for the development and validation of perception, state estimation, and control algorithms. Recent work has identified the \textit{sim2real} gap in optical tactile sensors, attributed to the unpredictable imperfections in real-world objects, with image augmentation suggested as a strategy for mitigation~\cite{Jianu2021Reducing}. In the context of LiDAR simulation, a \textit{``paired-scenario''} method has been proposed to evaluate and improve the simulation by creating digital twins of real-world scenarios, thus addressing the key factors in LiDAR modeling that contribute to the \textit{sim2real} gap~\cite{Manivasagam2023LiDARSim2Real}. Although, our approach also evaluates sim2real gaps using digital twins, we do not compare raw sensor signals. Instead, we judge the models by evaluating the performance of a downstream task that actually uses the raw sensor signals. For camera simulation, a \textit{``context-based''} approach is commonly employed to quantify the \textit{sim2real} gap, with style transfer techniques applied to bridge this discrepancy~\cite{elmquist_modeling_2021,elmquist_performance_2022,richter_enhancing_2023}.

\noindent{\textbf{\textit{Sim2Real} Gap in IMU/GPS Sensor Simulation}}. Despite recent advances in modeling IMU/GPS systems, there are few efforts to quantify and mitigate the \textit{sim2real} gap -- particularly in comparison to visual sensors. Hafez et al.~\cite{Hafez2012GPSNoiseDeduction} introduced a method to estimate the measurement noise covariance matrices, directly comparing simulated covariance with real-world data. The sensor model outputs were, however, not used in downstream robotics tasks, thus lacking a quantitative evaluation of how simulated covariance discrepancies translate to downstream task errors. Alternatively, Wang et al.~\cite{Wang2008GPSModelling} focused on modeling a GPS/INS navigation system, with less emphasis on validating the model and measuring the \textit{sim2real} gap. Although GPS/IMU sensors have been modeled across various simulators~\cite{airsim2018, asherSensorSimulation2021, KoenigDesign2004, carlaAVsim2017}, to the best of our knowledge, no methodology is in place for quantifying the \textit{sim2real} gap in IMU and GPS sensor simulations, especially in the context of downstream robotics tasks.

\subsection{Contribution}
From a high vantage point, in the context of velocity state estimation, we establish a methodology to quantitatively measure the \textit{sim2real} gap associated with IMU and GPS models. Drawing inspiration from~\cite{elmquist_performance_2022} and~\cite{Manivasagam2023LiDARSim2Real}, we employ the \textit{robot\_localization}~\cite{MooreStouchKeneralizedEkf2014} package as a \textit{``judge''} to evaluate the quality of the simulated sensor data. Our contributions are threefold. First, we introduce a novel, application specific methodology for evaluating the \textit{sim2real} gap in IMU and GPS sensor simulation. We show that the proposed metric is sharp as it can pinpoint the sim2real gap arising from the IMU / GPS sensor models. To our knowledge, this is the first attempt to quantify the \textit{sim2real} gap for this specific sensor modality. Second, we apply the proposed methodology to validate five different sensor models with 40 real-world tests, providing a comprehensive benchmark of the performance of various GPS and IMU sensor models for velocity estimation. We identify key aspects of GPS sensor models that enable small sim2real gaps. Lastly, we established a public dataset~\footnote[3]{\textbf{Dataset:}\;\url{https://shorturl.at/yBDKP}} and made available the open-source code~\footnote[4]{\textbf{Code:}\;\url{https://shorturl.at/nCFY1}} used in this investigation.

\section{METHODOLOGY}
\label{sec:method}
The \textit{Velocity Estimation Performance Difference (VEPD)} methodology proposed has five components described in subsections \S\ref{sec:method:judge} through \ref{sec:method:metrics}.

\subsection{Judge}
\label{sec:method:judge}
Our validation method indirectly assesses IMU and GPS sensor model fidelity by evaluating the performance of a downstream state estimation algorithm. We employ the widely used \textit{robot\_localization} package~\cite{MooreStouchKeneralizedEkf2014} to process IMU and GPS data, producing velocity estimates at 35 Hz. This package acts as a ``judge'' within our framework: the difference in its performance between simulation and reality is an indicator of how well a sensor model in simulation can mimic its real-world counterpart. By comparing the state estimator's output using simulated versus real sensor data, we can gauge the accuracy of our simulation's sensor models within the specific application of velocity state estimation.

\subsection{Real World Experiments}
\label{sec:method:reality}
\begin{figure}
    \centering
    \includegraphics[width=0.3\textwidth]{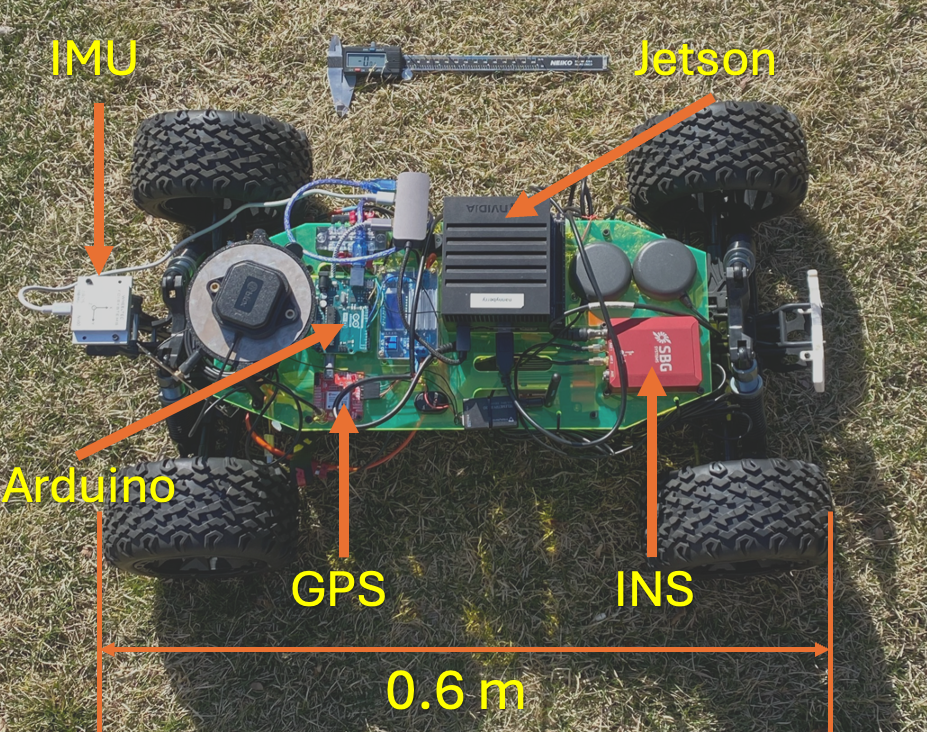}
    \caption{\textit{ART} -- the 1/6 scale vehicle (see 0.2 m vernier caliper for scale) used for real-world experiments.}
    \label{fig:art}
    \vspace{-10pt}
\end{figure}

We use the Autonomy Research Testbed (\textit{ART}) vehicle (1/6 scale)~\cite{artatkResearchPlatform2022}, equipped with a sensor suite (see Fig~\ref{fig:art}). Key components: IMU -- \textit{Wheeltec N-100 IMU} operating at 100 Hz for high-frequency inertial measurements; GPS -- \textit{Sparkfun ZED-F9P GPS-RTK2} operating at 10 Hz for centimeter-level positional accuracy ($\pm3$ cm) when paired with a base station; Compute -- \textit{Nvidia Jetson AGX Xavier} executing a ROS2 based autonomy stack~\cite{artatkResearchPlatform2022}. 

\subsection{Simulation Environment and Sensor Models}
\label{sec:method:simulation}
We utilize several IMU and GPS sensor models from two different simulators -- Chrono and AirSim.

\noindent{\textbf{Project Chrono}}. Chrono~\cite{chronoOverview2016} offers a high-fidelity, open-source simulation engine that is well-suited for simulating ground vehicles, providing a comprehensive suite of vehicle dynamics and sensor models. We create a digital-twin of the \textit{ART} vehicle within Chrono, replicating its physical properties and sensor suite. The IMU model utilizes a Gaussian drift noise model for both the gyroscope and accelerometer (see Appendix~\ref{sec:app:imu_gps_models} for details). We use two different GPS noise models in Chrono: one with an additive Gaussian noise, referred henceforth as \textit{Ch:Gauss} and the other with a modification of a random walk model, referred henceforth as \textit{Ch:RW}. The Chrono GPS sensor models do not model the GPS measurement noise covariance matrix; hence a zero matrix is passed to the state estimator. Appendix~\ref{sec:app:imu_gps_models} details both models along with parameter choices.

\noindent{\textbf{AirSim}}. AirSim is an open-source, cross-platform autonomous agent simulator built on Unreal Engine~\cite{airsim2018}. The IMU model is identical to the Gaussian drift noise model used in Chrono. The GPS sensor model is different in two ways: it models the Horizontal Dilution Of Precision (HDOP) used to model the measurement noise covariance; and it does not model GPS latitude/longitude noise. The HDOP is a measure of the geometric quality of a GPS satellite configuration in the sky~\cite{Jwo2001GPSHDOP,Zhong2006Geometric}. In AirSim, it is modeled using a low pass filter~\cite{airsim2018} (see Appendix~\ref{sec:app:airsim_imu_gps_models}).

To enable the use of the existing Chrono digital twin of ART~\cite{huzaifaIEEE-AccessCalibration2024}, we implement the AirSim GPS and IMU models in Project Chrono and use its ROS2 bridge, Chrono::ROS, for communication. Additionally, the simulated GPS and IMU data is configured to publish at 10 Hz and 100 Hz, respectively, mirroring our real-world sensor setup for all combinations of sensor models. Chrono adds noise directly to ground-truth latitude/longitude measurements while setting the HDOP to zero. In contrast, AirSim models measurement noise covariance through HDOP but adds no noise to ground-truth latitude/longitude measurements.

\subsection{Ground Truth Velocity}
\label{sec:method:ground_truth}
To evaluate the precision of the velocity estimation process, we require a ground truth. In simulation this is directly provided by querying the simulator. In real tests, we employ a high-end commercial \textit{SBG Ellipse2-D} Inertial Navigation System (INS) to produce ground truth (see Fig.~\ref{fig:art}). To validate the accuracy of the INS velocity, we conduct a straight-line experiment, measuring the vehicle's distance over 30 seconds with an RTK-GPS. This experiment is repeated 20 times to obtain a statistical measure. We then calculate the average speed (distance/time) and compare it to the INS-recorded speed. We observe an average error of 0.03 m/s, confirming the reliability of the INS system as a ground-truth velocity source for further validation. Note that for these experiments, the RTK-GPS is only used to measure distance, for which it is highly accurate.

\subsection{Error Metric}
\label{sec:method:metrics}
The Velocity Estimation Performance Difference (VEPD) metric provides a comprehensive evaluation of the sim2real gap in GPS/IMU sensor simulations by comparing velocity error \textit{signatures} between estimated and ground truth velocity in simulation and reality. For each experiment, we obtain velocity profiles from the state estimator, both when using real-world sensor data ($\vv_{re} \in \mathbb{R}^T$) and simulated data ($\vv_{se} \in \mathbb{R}^T$).  We also obtain ground-truth velocities from the simulators ($\vv_{sg} \in \mathbb{R}^T$) and the real-world INS system ($\vv_{rg} \in \mathbb{R}^T$).  The Root Mean Squared Errors (RMSE) for both simulation and real-world scenarios is calculated as
\begin{equation}
    E_{\alpha} = \sqrt{\frac{1}{T} \sum_{t=1}^{T} (\vv_{\alpha e}[t] - \vv_{\alpha g}[t])^2} \; ,
\label{eq:RMSE}
\end{equation}
where $E_r$ and $E_s$ are obtained by replacing $\alpha$ with $r$ and $s$, respectively, and $T$ is the number of time steps in the trajectory. We perform $K$ tests to obtain $K$ RMSE values for reality and simulation. This provides us with two point sets of $K$ points each,  $A_1 = \{E_{r}^1, E_{r}^2, \ldots, E_{r}^K\} \in \mathbb{R}^K$ and $B_1 = \{E_{s}^1, E_{s}^2, \ldots, E_{s}^K\} \in \mathbb{R}^K$. The Wasserstein distance, given by $W_1(A_1, B_1) = \inf_{\gamma \in \Gamma(A_1, B_1)} \int_{X \times Y} \|x - y\| d\gamma(x, y) \in [0, 1]$, is then used to compare the two distributions to understand how closely simulation velocity errors align with real-world velocity errors~\cite{Villani2009}. For a small sim2real gap, we would like $W_1$ to be small, i.e. the RMSE between the velocity estimate in sim and the ground truth velocity in sim should be similar to the RMSE between velocity estimate in reality and the ground truth velocity in reality.  

The RMSE, however, does not capture the \textit{signature} of the velocity estimate signal, meaning that two signals with identical RMSE can exhibit substantially different patterns over time. To analyze error patterns, we first compute the Fast Fourier Transform (FFT) of the velocity signals and use the Weiner entropy $S$ to categorize how noise-like a signal is. A higher entropy value indicates that the energy distribution across the spectrum is more uniform, which typically corresponds to noise, whereas a lower entropy value indicates the presence of prominent peaks in the spectrum, characteristic of tonal signals. For the FFT transformed signal $\vv^f_{se} = FFT(\vv_{se})$, $S \in \mathbb{R}$ is given by
\begin{align}
    S_{se} = \frac{\exp\left(\frac{1}{N} \sum_{i=0}^{N-1} \ln |v^f_{se}(i)|\right)}{\frac{1}{N} \sum_{i=0}^{N-1} |v^f_{se}(i)|} \; .
    \label{eq:ent}
\end{align}
To compare the entropy of the simulated velocity estimate against the real-world estimate, while accounting for differences from their respective ground truths, we compute the absolute difference between them:
\begin{equation}
    \begin{split}
    S_s = |S_{se} - S_{sg}| \\
    S_r = |S_{re} - S_{rg}| 
    \end{split} \quad .
\label{eq:ent_diff}
\end{equation}
With $K$ tests we end up with two point sets, $A_2 = \{S_{r}^1, S_{r}^2, \ldots, S_{r}^K\}$ and  $B_2 = \{S_{s}^1, S_{s}^2, \ldots, S_{s}^K\}$, and we again compute the Wasserstein distance $W_2(A_2,B_2)$.  

Finally, the VEPD metric is the average of the individual Wasserstein distances -- a measure of the difference in the velocity error signatures in sim and real:
\begin{align}
    \text{VEPD} = \frac{W_1 + W_2}{2} \in [0, 1] \; .
\end{align}

VEPD provides a balanced view as it considers both the accuracy and quality of the velocity signals. By considering distribution shapes and signal characteristics, VEPD goes beyond point-by-point comparisons, helping it provide more context. Figure~\ref{fig:flow_chart} visually summarizes the methodological approach for evaluating the sim2real gap in GPS/IMU sensor simulation.

\section{Results}
\label{sec:exp}
\begin{figure*}
    \centering
    \subfloat[Histograms of Root Mean Squared Error (RMSE) (see Eq.~\ref{eq:RMSE})]{\includegraphics[width=1.0\textwidth]{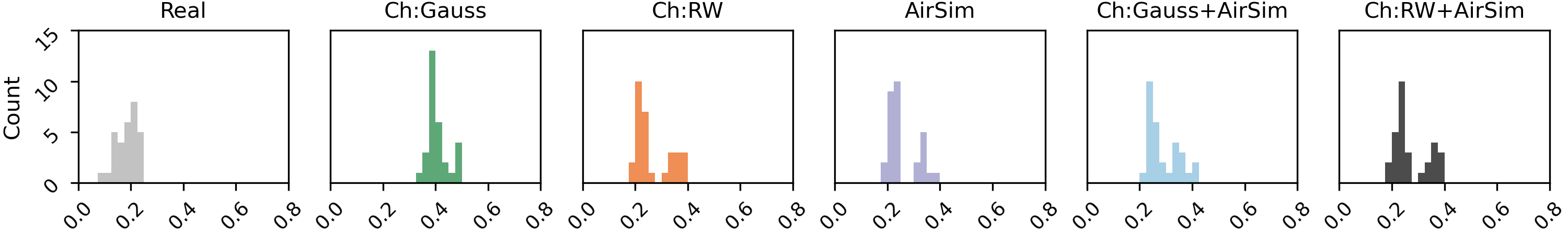}\label{fig:hist_rmse}}
    \par
    \subfloat[Histograms of Wiener entropy's (see Eq.~\ref{eq:ent})]{\includegraphics[width=1.0\textwidth]{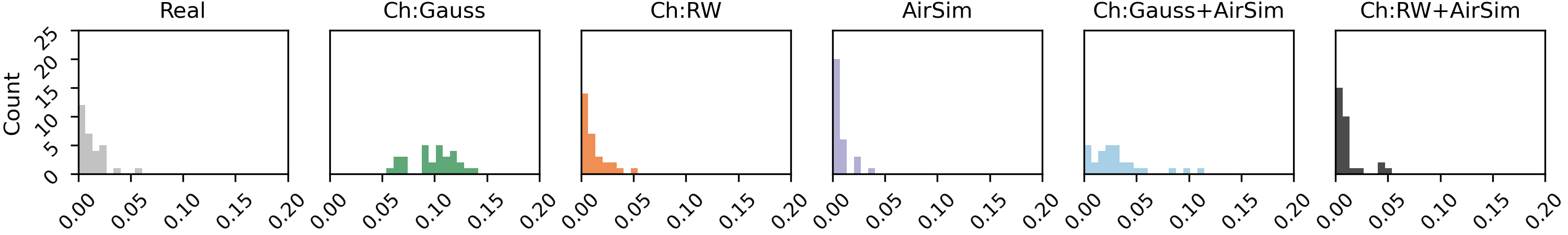}\label{fig:hist_fft}}
    \caption{Comparison of the distributions of RMSE and Wiener entropy for velocity estimates obtained from different simulated GPS sensor models. Lower RMSE and Wiener entropy values indicate a closer match between simulation and reality. The Ch:RW model (RW stands for Random Walk) demonstrates the most similar Wiener Entropy distribution to the real-world data. Conversely, the AirSim model, with minimal modeled noise, achieves the closest RMSE distribution to reality.}

\end{figure*}
Experiments are conducted to answer the following:
\begin{enumerate}
    \item[A.] How big is the sim2real gap for the IMU sensor, when it is singled out? 
    \item[B.] What aspects of GPS modeling are most important to consider for velocity state estimation?
    \item[C.] Does the type of vehicle maneuver (e.g., straight-line, half-sine curve, circle) affect the VEPD scores for the simulated sensors?
    \item[D.] Is VEPD agnostic to other sim2real differences arising from say, the dynamics of the vehicle or the environment?
\end{enumerate}

\smallskip

\begin{figure}
    \centering
    \includegraphics[width=0.4\textwidth]{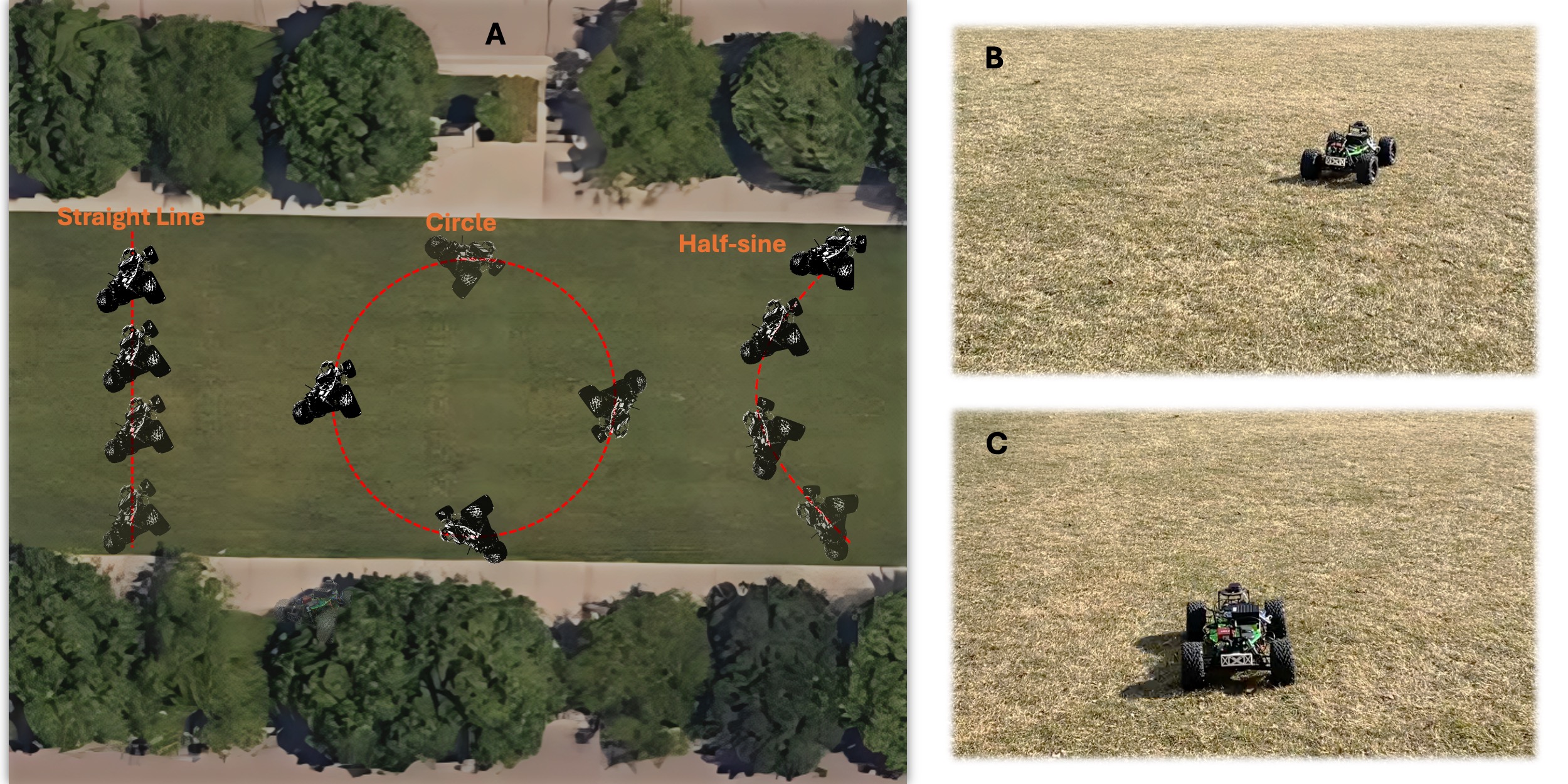}
    \caption{\textbf{A.} The three dynamic tests performed in the real-world on a patch of grass. \textbf{B. and C.} Snapshots of the vehicle from a video feed during the half-sine curve test.} 
    \label{fig:outdoor_tests}
\end{figure}

To address the posed questions, we generate a dataset with three different vehicle trajectories: straight line, half-sine curve, and circle in both simulated and real-world environments (see Fig.~\ref{fig:outdoor_tests}). Each scenario is tested 10 times, with the reality tests performed on off-road terrain (grass field) and simulation tests on flat terrain. We utilize a waypoint-based neural network (NN) controller to navigate the vehicle through both simulated and real-world environments. Using our methodology (Sec.~\ref{sec:method}, Fig.~\ref{fig:flow_chart}), we calculate the VEPD metric for each scenario. 

\subsection{Sim2Real gap for the IMU sensor}
\label{sec:exp:imu_noise}
\noindent {\textbf{Approach:}} To investigate the contribution of the simulated IMU sensor to the sim2real gap in velocity estimation, we use 10 reality tests from the straight-line scenario to obtain one set of velocity estimates and ground truth velocities (termed \textit{Real IMU + Real GPS}). Additionally, the GPS data, collected as ROS2 bags from each of the reality tests, is played in simulation alongside the simulated IMU sensor, through the state-estimator, to obtain another set of velocity estimates and ground truth velocities (termed \textit{Sim. IMU + Real GPS}). The velocity error signatures \textit{Real IMU + Real GPS} and \textit{Sim. IMU + Real GPS} are then compared by computing the VEPD score using the methodology described earlier (Sec.~\ref{sec:method:metrics}). This setup enables us to isolate the influence of the simulated IMU on the velocity estimate.

\smallskip

\begin{figure}
    \centering
    \includegraphics[width=0.5\textwidth]{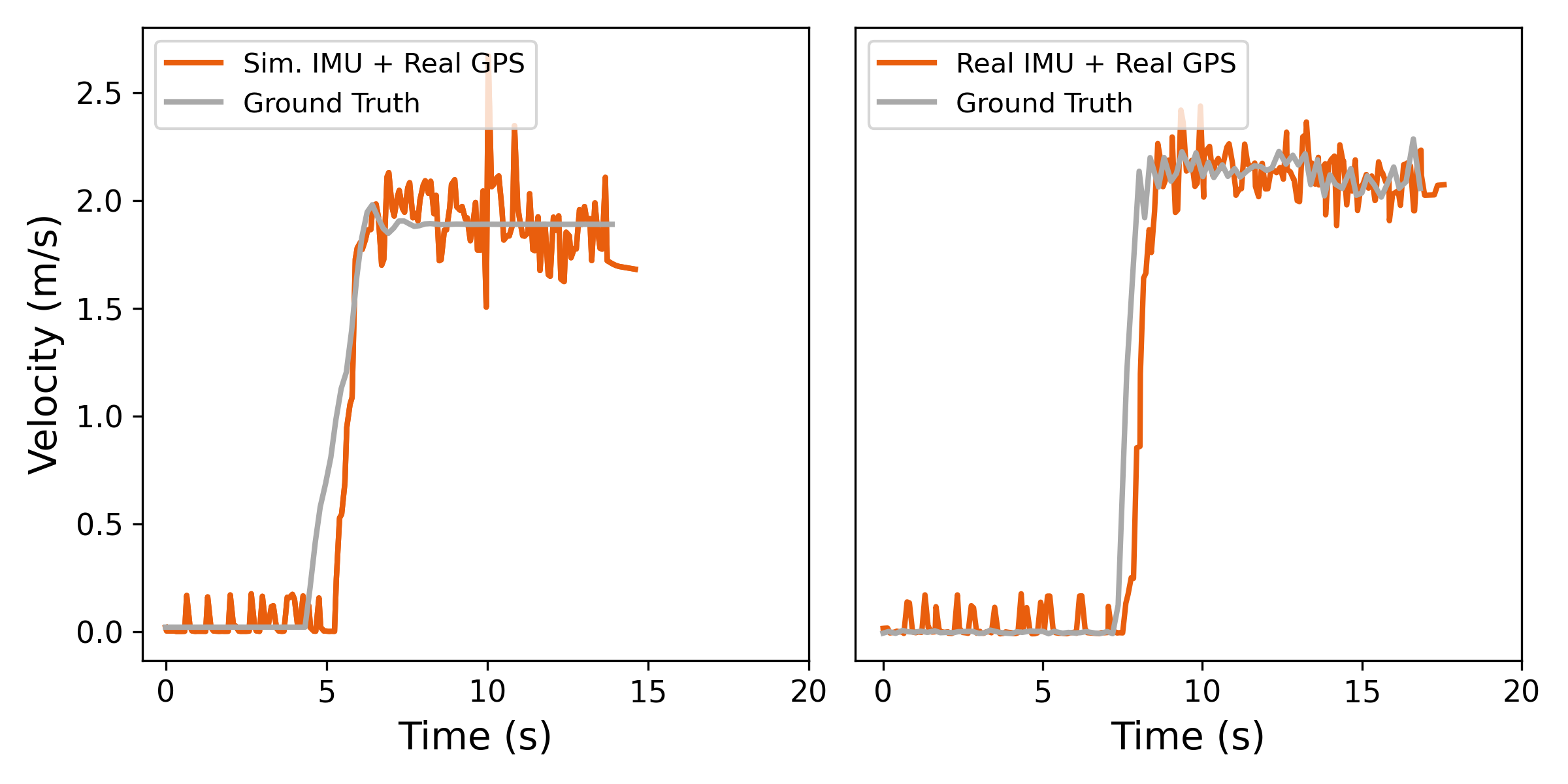}
    \caption{Using the simulated IMU with real GPS produces a velocity estimate (left orange) that closely resembles the setup when a real IMU is used with a real GPS (right orange). The RMSE (Left: $0.2004$, Right: $0.2083$) and Wiener Entropy (Left: $0.0311$, Right: $0.0272$) are close, which yield a small VEPD.} 
    \label{fig:imu_not_needed}
\end{figure}

\noindent {\textbf{Discussion:}} We obtain a VEPD score of $0.0211$, which suggests that the distributions of RMSEs and Wiener Entropy differences between the \textit{Real IMU + Real GPS} and the \textit{Sim. IMU + Real GPS} are close. To gain an intuitive understanding of the score, we plot the velocities from one pair of tests in Fig.~\ref{fig:imu_not_needed} and observe that the velocity signatures are similar.

\subsection{What aspects are important for GPS modeling?}
\label{sec:exp:important_gps}
\noindent {\textbf{Approach:}} For analyzing the GPS sensor models in isolation, we use results from Sec.~\ref{sec:exp:imu_noise} to fix the IMU sensor model and its parameters across all tests. There are two important aspects of GPS modeling that we consider: 1) the noise model applied to the latitude and longitude, and 2) the measurement covariance matrix of the GPS sensor, which is governed by the Dilution Of Precision (DOP) values. To understand the importance of these aspects, we simulate the GPS sensor in five different variants: 1) with additive Gaussian noise and zero measurement covariance (Ch:Gauss); 2) with Random Walk noise and zero measurement covariance (Ch:RW); 3) with no noise and measurement covariance given by Eq.~\ref{eq:airsim_cov} (AirSim); 4) with additive Gaussian noise and measurement covariance given by Eq.~\ref{eq:airsim_cov} (Ch:Gauss+AirSim); and 5) with Random Walk noise and measurement covariance given by Eq.~\ref{eq:airsim_cov} (Ch:RW+AirSim). For more details about the noise models and the parameters used, see Appendix~\ref{sec:app:airsim_imu_gps_models}. The same methodology discussed in Sec.~\ref{sec:method:metrics} is used with $K=30$ tests, with 10 tests in each dynamic scenario. These 30 tests are run for each of the five sensor noise variants to provide 150 simulation tests. The VEPD metric is evaluated between each sensor variant and reality (Table.~\ref{tab:vepd_combined}), and histograms of the RMSE and Wiener Entropy differences are plotted (Fig.~\ref{fig:hist_rmse},~\ref{fig:hist_fft}).

\smallskip

\noindent {\textbf{Discussion:}} Figure~\ref{fig:hist_fft} shows how ``noise-like'' the simulated velocity estimates are compared to real-world data. The Ch:RW GPS model's Wiener Entropy distribution most closely resembles the ``real'' case,  confirmed by its lowest $W_2$ value (Table.~\ref{tab:vepd_combined}). Interestingly, the Ch:Gauss GPS model (based on additive Gaussian noise) performs poorly without measurement covariance but improves 8x when covariance is modeled.  This suggests the state estimator likely benefits from assigning lower weights to noisy Ch:Gauss outputs based on high initial covariance values (see Eq.~\ref{eq:airsim_cov}). Conversely, this explains why modeling measurement covariance slightly worsens Ch:RW performance, as its random-walk-based initialization is closer to the ground truth, and higher covariance values may mislead the estimator.
Fig.~\ref{fig:hist_rmse} and Table~\ref{tab:vepd_combined} demonstrate that the AirSim GPS model yields the smallest difference in velocity error between simulation and reality, as measured by RMSE (Eq.~\ref{eq:RMSE}), followed closely by Ch:RW.  We hypothesize that AirSim's lack of explicit latitude/longitude noise modeling contributes to its strong performance. This aligns with the low-noise characteristics of the real-world RTK-GPS used in our experiments, which has a 3 cm accuracy.
The analysis suggests that for GPS sensors, accurately modeling measurement noise covariance is crucial for achieving realistic velocity estimates in simulation, especially when sensor noise characteristics resemble those from additive Gaussian models.  Furthermore, the Ch:RW model's random-walk nature appears to effectively capture the noise characteristics observed in reality. When real-world sensor noise is minimal (as with high-precision RTK-GPS), minimizing additional simulated latitude/longitude noise can further improve simulation fidelity.
\begin{table}
    \centering
    \caption{VEPD values for the sensor model variants. Low values are indicative of good performance.}
    \label{tab:vepd_combined}
    \begin{tabular}{lccc}
        \toprule
        Sensor model & $W_1$ & $W_2$ & VEPD \\ 
        \midrule
        Ch:Gauss&0.2242&0.0859&0.155\\
        Ch:RW &0.0836 & \textbf{0.0025} & 0.043\\
        AirSim&\textbf{0.0725}&0.0054&\textbf{0.039}\\
        Ch:Gauss+AirSim&0.1069&0.0185&0.0627\\
        Ch:RW+AirSim&0.092&0.0044&0.0482\\
        \bottomrule
    \end{tabular}
\end{table}
\subsection{Does the type of vehicle maneuver affect the VEPD scores for the simulated sensors?}
\noindent {\textbf{Approach:}} To investigate the impact of maneuvers on VEPD, we calculate the metric on a scenario-by-scenario basis. This approach yields three VEPD values per sensor variant, one for each test scenario. Table~\ref{tab:vepd_scenario} summarizes these values and assigns ranks (lower is better) to each sensor variant within a specific scenario.  

\smallskip

\noindent {\textbf{Discussion:}} Ideally, sensor variant rankings should remain consistent across scenarios, as dynamic maneuvers should not alter metrics that evaluate sensor model performance.  While Table~\ref{tab:vepd_scenario} shows some minor variations in ranking, the overall trends hold: Ch:RW and AirSim consistently perform well, while Ch:Gauss ranks lower. 

\begin{table}
    \centering
    \caption{VEPD for each sensor variant across different dynamic scenarios. R stands for rank.}
    \label{tab:vepd_scenario}
    \begin{tabular}{lcccccc}
        \toprule
        & \multicolumn{2}{c}{Line} & \multicolumn{2}{c}{Circle} & \multicolumn{2}{c}{Sine} \\
        \cmidrule(lr){2-3} \cmidrule(lr){4-5} \cmidrule(lr){6-7}
        Sensor model & VEPD & R & VEPD & R & VEPD & R \\ 
        \midrule
        Ch:Gauss & 0.1387 & 5 & 0.1763 & 5 & 0.1501 & 5\\
        Ch:RW & \textbf{0.0048} & \textbf{1} &0.0488& 2 & 0.0796 & 2\\
        AirSim & 0.0066& 2 & \textbf{0.0459} & \textbf{1} & \textbf{0.066} & \textbf{1}\\
        Ch:Gauss+AirSim&0.026& 4 & 0.0842& 4 & 0.0796 & 3\\
        Ch:RW+AirSim&0.0085& 3 &0.0574& 3 & 0.0799 & 4\\
        \bottomrule
    \end{tabular}
\end{table}

\subsection{Is \textit{VEPD} agnostic to other sim2real differences such as varying robot dynamics?}
\noindent {\textbf{Approach:}} To explore the robustness of VEPD against environmental and dynamic differences between sim and real, we perform the following study. We run 10 sine scenarios on inclined concrete in reality, complementing the original grass surface tests. The same 10 sine scenarios from Sec.~\ref{sec:exp:important_gps} performed in simulation on flat, rigid terrain are then used to calculate two VEPD values: (i) \textit{sim2grass}: compares the simulated sine tests to the real-world grass tests; (ii) \textit{sim2concrete}: compares the simulated sine tests to the real-world concrete tests. We pick Ch:RW, which a good performer.

\smallskip

\begin{figure}
    \centering
    \includegraphics[width=0.48\textwidth]{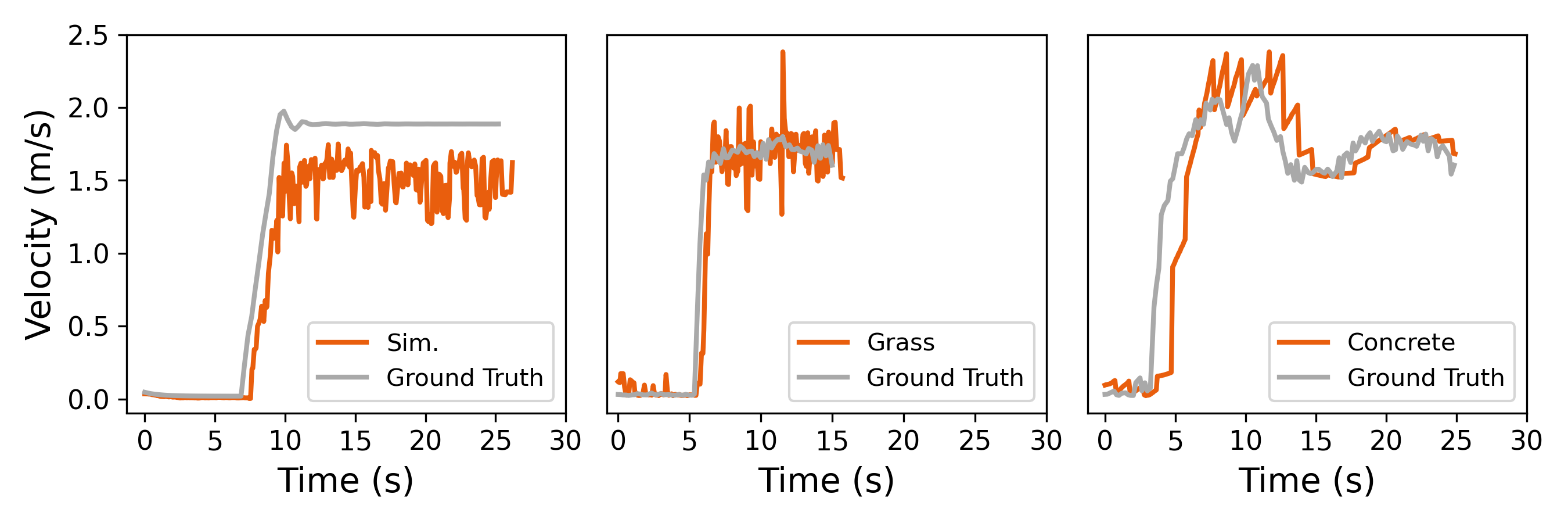}
    \caption{The velocity estimates (orange) produced by the state estimator across three varying environments for one out of the ten tests. \textbf{Left:} In simulation with the Ch:RW GPS noise model, \textbf{Center:} In reality using real sensors on grass, \textbf{Right:} In reality using real sensors on inclined concrete. Although the plots show varying velocity profiles due to varying environments, the VEPD score of the sensor model remains the same.} 
    \label{fig:agnostic}
\end{figure}

\noindent {\textbf{Discussion:}} We observe a VEPD of 0.0796 for sim2grass and 0.0867 for sim2concrete ($\approx 8\%$ difference). This similarity in VEPD values, despite environmental changes, suggests that the metric remains consistent across sim2real differences not directly related to GPS/IMU sensor modeling.  Figure~\ref{fig:agnostic} further supports this finding, visually demonstrating how distinct vehicle dynamics across the tested environments do not significantly impact the VEPD. This agnostic nature is valuable -- VEPD allows for focused evaluation of sensor models without confounding factors due to other differences between the simulation and real-world environment.

\section{CONCLUSION}
\label{sec:conc}
This work establishes a framework to systematically evaluate the realism of sensor simulation in robotics applications.
We introduce the Velocity Estimation Performance Difference (VEPD), a metric tailored for quantifying the sim2real gap in GPS/IMU sensor modeling for velocity state estimation. This metric is used to benchmark five GPS sensor model variants across 40 real-world tests to understand the influence of specific GPS modeling assumptions on the estimated velocity. Our experiments highlight the important role of measurement noise covariance in achieving realistic GPS simulations. The Ch:RW GPS model based on a random walk often demonstrated noise patterns closely resembling real-world data. The VEPD metric proved robust to changes in dynamics and environment, enabling focused evaluation of sensor model realism.   

Future work includes evaluating the robustness of VEPD across different state-estimators to understand how much the choice of the ``judge'' influences the score provided to the sensor model. Additionally, it remains to explore velocity estimation in GPS-constrained or high GPS noise environments where different features of GPS models could become a dominant factor in velocity estimation. While the focus is on GPS/IMU sensors, the core concepts behind VEPD could potentially be adapted to assess the sim2real gap in other sensor models. Finally, we plan to include CARLA and Gazebo in a future study to gain a better insight into the accuracy of the GPS/IMU sensors used therein.

%%%%%%%%%%%%%%%%%%%%%%%%%%%%%%%%%%%%%%%%%%%%%%%%%%%%%%%%%%%%%%%%%%%%%%%%%%%%%%%%

%%%%%%%%%%%%%%%%%%%%%%%%%%%%%%%%%%%%%%%%%%%%%%%%%%%%%%%%%%%%%%%%%%%%%%%%%%%%%%%%

\section*{ACKNOWLEDGMENT}
This work was in part made possible with funding from NSF project CMMI215385.

% \clearpage

\bibliographystyle{IEEEtran}
\bibliography{BibFiles/refsAutonomousVehicles,BibFiles/refsTerramech,BibFiles/refsChronoSpecific,BibFiles/refsRobotics,BibFiles/refsSBELspecific,BibFiles/refsMachineLearning,BibFiles/refsMLPhysics,BibFiles/refsMBS,BibFiles/refsCompSci,BibFiles/refsStatsML,BibFiles/refsFSI,BibFiles/refsDEM,BibFiles/refsOddsEnds, BibFiles/refsSensors, BibFiles/refsTrafficSim, BibFiles/refsSimToRealGapCamera}

% Generated by IEEEtran.bst, version: 1.14 (2015/08/26)
\def\cprime{$'$}
\begin{thebibliography}{10}
\providecommand{\url}[1]{#1}
\csname url@samestyle\endcsname
\providecommand{\newblock}{\relax}
\providecommand{\bibinfo}[2]{#2}
\providecommand{\BIBentrySTDinterwordspacing}{\spaceskip=0pt\relax}
\providecommand{\BIBentryALTinterwordstretchfactor}{4}
\providecommand{\BIBentryALTinterwordspacing}{\spaceskip=\fontdimen2\font plus
\BIBentryALTinterwordstretchfactor\fontdimen3\font minus
  \fontdimen4\font\relax}
\providecommand{\BIBforeignlanguage}[2]{{%
\expandafter\ifx\csname l@#1\endcsname\relax
\typeout{** WARNING: IEEEtran.bst: No hyphenation pattern has been}%
\typeout{** loaded for the language `#1'. Using the pattern for}%
\typeout{** the default language instead.}%
\else
\language=\csname l@#1\endcsname
\fi
#2}}
\providecommand{\BIBdecl}{\relax}
\BIBdecl

\bibitem{Ding2020LongitudinalVS}
\BIBentryALTinterwordspacing
X.~Ding, Z.~Wang, L.~Zhang, and C.~Wang, ``Longitudinal vehicle speed
  estimation for four-wheel-independently-actuated electric vehicles based on
  multi-sensor fusion,'' \emph{IEEE Transactions on Vehicular Technology},
  vol.~69, pp. 12\,797--12\,806, 2020. [Online]. Available:
  \url{https://api.semanticscholar.org/CorpusID:226663718}
\BIBentrySTDinterwordspacing

\bibitem{Gu2020NonlinearOD}
\BIBentryALTinterwordspacing
N.~Gu, Z.~Peng, D.~Wang, L.~Liu, and Y.~Jiang, ``Nonlinear observer design for
  a robotic unmanned surface vehicle with experiment results,'' \emph{Applied
  Ocean Research}, vol.~95, p. 102028, 2020. [Online]. Available:
  \url{https://api.semanticscholar.org/CorpusID:214055243}
\BIBentrySTDinterwordspacing

\bibitem{Sukkarieh1999HighIntegrityIMU}
S.~Sukkarieh, E.~Nebot, and H.~Durrant-Whyte, ``A high integrity imu/gps
  navigation loop for autonomous land vehicle applications,'' \emph{IEEE
  Transactions on Robotics and Automation}, vol.~15, no.~3, pp. 572--578, 1999.

\bibitem{Saeedmanesh2021}
\BIBentryALTinterwordspacing
M.~Saeedmanesh, A.~Kouvelas, and N.~Geroliminis, ``An extended {K}alman filter
  approach for real-time state estimation in multi-region {MFD} urban
  networks,'' \emph{Transportation Research Part C: Emerging Technologies},
  vol. 132, p. 103384, 2021. [Online]. Available:
  \url{https://www.sciencedirect.com/science/article/pii/S0968090X21003831}
\BIBentrySTDinterwordspacing

\bibitem{Jaradat2017NonLinearAD}
\BIBentryALTinterwordspacing
M.~A.~K. Jaradat and M.~F. Abdel-Hafez, ``Non-linear autoregressive
  delay-dependent {INS/GPS} navigation system using neural networks,''
  \emph{IEEE Sensors Journal}, vol.~17, pp. 1105--1115, 2017. [Online].
  Available: \url{https://api.semanticscholar.org/CorpusID:23079458}
\BIBentrySTDinterwordspacing

\bibitem{kadian2020sim2real}
A.~Kadian, J.~Truong, A.~Gokaslan, A.~Clegg, E.~Wijmans, S.~Lee, M.~Savva,
  S.~Chernova, and D.~Batra, ``{Sim2Real} predictivity: Does evaluation in
  simulation predict real-world performance?'' \emph{IEEE Robotics and
  Automation Letters}, vol.~5, no.~4, pp. 6670--6677, 2020.

\bibitem{Manivasagam2023LiDARSim2Real}
S.~Manivasagam, I.~A. B\^arsan, J.~Wang, Z.~Yang, and R.~Urtasun, ``Towards
  zero domain gap: A comprehensive study of realistic lidar simulation for
  autonomy testing,'' in \emph{Proceedings of the IEEE/CVF International
  Conference on Computer Vision (ICCV)}, October 2023, pp. 8272--8282.

\bibitem{Wang2008GPSModelling}
L.~Wang, H.~Zhao, and X.~Yang, ``The modeling and simulation for {GPS/INS}
  integrated navigation system,'' in \emph{2008 International Conference on
  Microwave and Millimeter Wave Technology}, vol.~4, 2008, pp. 1991--1994.

\bibitem{PNASsimRobotics2021}
\BIBentryALTinterwordspacing
H.~Choi, C.~Crump, C.~Duriez, A.~Elmquist, G.~Hager, D.~Han, F.~Hearl,
  J.~Hodgins, A.~Jain, F.~Leve, C.~Li, F.~Meier, D.~Negrut, L.~Righetti,
  A.~Rodriguez, J.~Tan, and J.~Trinkle, ``On the use of simulation in robotics:
  Opportunities, challenges, and suggestions for moving forward,''
  \emph{{Proceedings of the National Academy of Sciences}}, vol. 118, no.~1,
  2021. [Online]. Available:
  \url{https://www.pnas.org/content/118/1/e1907856118}
\BIBentrySTDinterwordspacing

\bibitem{karenSimRobotics2021}
\BIBentryALTinterwordspacing
K.~Liu and D.~Negrut, ``The role of physics-based simulators in robotics,''
  \emph{Annual Review of Control, Robotics, and Autonomous Systems}, vol.~4,
  no.~1, pp. 35--58, 2021. [Online]. Available:
  \url{https://doi.org/10.1146/annurev-control-072220-093055}
\BIBentrySTDinterwordspacing

\bibitem{Xuemin2023Sim2RealSurvey}
X.~Hu, S.~Li, T.~Huang, B.~Tang, R.~Huai, and L.~Chen, ``How simulation helps
  autonomous driving: A survey of sim2real, digital twins, and parallel
  intelligence,'' \emph{IEEE Transactions on Intelligent Vehicles}, pp. 1--20,
  2023.

\bibitem{todorovComparisonSimEngines2015}
T.~Erez, Y.~Tassa, and E.~Todorov, ``Simulation tools for model-based robotics:
  Comparison of {Bullet}, {Havok}, {MuJoCo}, {ODE} and {PhysX},'' in \emph{2015
  IEEE international conference on robotics and automation (ICRA)}.\hskip 1em
  plus 0.5em minus 0.4em\relax IEEE, 2015, pp. 4397--4404.

\bibitem{KoenigDesign2004}
N.~Koenig and A.~Howard, ``Design and use paradigms for {Gazebo}, an
  open-source multi-robot simulator,'' in \emph{{IEEE}/{RSJ} {International}
  {Conference} on {Intelligent} {Robots} and {Systems} ({IROS})}, vol.~3,
  Sendai, Japan, Sep. 2004, pp. 2149--2154.

\bibitem{mujoco2012}
E.~Todorov, T.~Erez, and Y.~Tassa, ``{MuJoCo}: A physics engine for model-based
  control,'' in \emph{2012 IEEE/RSJ International Conference on Intelligent
  Robots and Systems}, 2012, pp. 5026--5033.

\bibitem{Pybullet2023}
\BIBentryALTinterwordspacing
E.~Coumans and Y.~Bai, ``{PyBullet}: A {P}ython module for physics
  simulation,'' 2023, available: \url{http://pybullet.org}. [Online].
  Available: \url{http://pybullet.org}
\BIBentrySTDinterwordspacing

\bibitem{IssacNvidia2023}
\BIBentryALTinterwordspacing
{NVIDIA Corporation}, ``{Isaac SDK: NVIDIA's Robotics Platform},'' 2023,
  available: \url{https://developer.nvidia.com/isaac-sdk}. [Online]. Available:
  \url{https://developer.nvidia.com/isaac-sdk}
\BIBentrySTDinterwordspacing

\bibitem{chronoOverview2016}
A.~Tasora, R.~Serban, H.~Mazhar, A.~Pazouki, D.~Melanz, J.~Fleischmann,
  M.~Taylor, H.~Sugiyama, and D.~Negrut, ``{Chrono}: An open source
  multi-physics dynamics engine,'' in \emph{High Performance Computing in
  Science and Engineering -- Lecture Notes in Computer Science}, T.~Kozubek,
  Ed.\hskip 1em plus 0.5em minus 0.4em\relax Springer International Publishing,
  2016, pp. 19--49.

\bibitem{Afzal2021SimulationRoboticsTestAutomation}
A.~Afzal, D.~S. Katz, C.~Le~Goues, and C.~S. Timperley, ``Simulation for
  robotics test automation: Developer perspectives,'' in \emph{2021 14th IEEE
  Conference on Software Testing, Verification and Validation (ICST)}, 2021,
  pp. 263--274.

\bibitem{Jianu2021Reducing}
T.~Jianu, D.~F. Gomes, and S.~Luo, ``Reducing tactile sim2real domain gaps via
  deep texture generation networks,'' \emph{2022 International Conference on
  Robotics and Automation (ICRA)}, pp. 8305--8311, 2021.

\bibitem{elmquist_modeling_2021}
A.~Elmquist and D.~Negrut, ``Modeling {Cameras} for {Autonomous} {Vehicle} and
  {Robot} {Simulation}: {An} {Overview},'' \emph{IEEE Sensors Journal},
  vol.~21, no.~22, pp. 25\,547--25\,560, 2021.

\bibitem{elmquist_performance_2022}
A.~Elmquist, R.~Serban, and D.~Negrut, ``A performance contextualization
  approach to validating camera models for robot simulation,'' \emph{arXiv
  e-prints}, p. arXiv:2208.01022, Aug. 2022.

\bibitem{richter_enhancing_2023}
S.~R. Richter, H.~A. Alhaija, and V.~Koltun, ``Enhancing {Photorealism}
  {Enhancement},'' \emph{IEEE Transactions on Pattern Analysis and Machine
  Intelligence}, vol.~45, no.~2, pp. 1700--1715, 2023.

\bibitem{Hafez2012GPSNoiseDeduction}
\BIBentryALTinterwordspacing
M.~F. Abdel-Hafez, ``On the {GPS/IMU} sensors' noise estimation for enhanced
  navigation integrity,'' \emph{Mathematics and Computers in Simulation},
  vol.~86, pp. 101--117, 2012, the Seventh International Symposium on Neural
  Networks + The Conference on Modelling and Optimization of Structures,
  Processes and Systems. [Online]. Available:
  \url{https://www.sciencedirect.com/science/article/pii/S0378475410000765}
\BIBentrySTDinterwordspacing

\bibitem{airsim2018}
S.~Shah, D.~Dey, C.~Lovett, and A.~Kapoor, ``{AirSim}: High-fidelity visual and
  physical simulation for autonomous vehicles,'' in \emph{Field and service
  robotics}.\hskip 1em plus 0.5em minus 0.4em\relax Springer, 2018, pp.
  621--635.

\bibitem{asherSensorSimulation2021}
A.~Elmquist, R.~Serban, and D.~Negrut, ``A sensor simulation framework for
  training and testing robots and autonomous vehicles,'' \emph{Journal of
  Autonomous Vehicles and Systems}, vol.~1, no.~2, p. 021001, 2021.

\bibitem{carlaAVsim2017}
A.~Dosovitskiy, G.~Ros, F.~Codevilla, A.~Lopez, and V.~Koltun, ``{CARLA}: {An}
  open urban driving simulator,'' in \emph{Proceedings of the 1st Annual
  Conference on Robot Learning}, 2017, pp. 1--16.

\bibitem{MooreStouchKeneralizedEkf2014}
T.~Moore and D.~Stouch, ``A generalized {Extended Kalman Filter} implementation
  for the robot operating system,'' in \emph{Proceedings of the 13th
  International Conference on Intelligent Autonomous Systems (IAS-13)}.\hskip
  1em plus 0.5em minus 0.4em\relax Springer, July 2014.

\bibitem{artatkResearchPlatform2022}
\BIBentryALTinterwordspacing
A.~Elmquist, A.~Young, T.~Hansen, S.~Ashokkumar, S.~Caldararu, A.~Dashora,
  I.~Mahajan, H.~Zhang, L.~Fang, H.~Shen, X.~Xu, R.~Serban, and D.~Negrut,
  ``{ART/ATK}: A research platform for assessing and mitigating the sim-to-real
  gap in robotics and autonomous vehicle engineering,'' 2022. [Online].
  Available: \url{https://arxiv.org/pdf/2211.04886.pdf}
\BIBentrySTDinterwordspacing

\bibitem{Jwo2001GPSHDOP}
D.-J. Jwo, ``Efficient {DOP} calculation for {GPS} with and without altimeter
  aiding,'' \emph{Journal of Navigation}, vol.~54, pp. 269 -- 279, 2001.

\bibitem{Zhong2006Geometric}
E.-J. Zhong and T.~Huang, ``Geometric dilution of precision in navigation
  computation,'' \emph{2006 International Conference on Machine Learning and
  Cybernetics}, pp. 4116--4119, 2006.

\bibitem{huzaifaIEEE-AccessCalibration2024}
H.~Unjhawala, T.~Hansen, H.~Zhang, S.~Caldraru, S.~Chatterjee, L.~Bakke, J.~Wu,
  R.~Serban, and D.~Negrut, ``An expeditious and expressive vehicle dynamics
  model for applications in controls and reinforcement learning,'' \emph{IEEE
  Access}, pp. 1--1, 2024.

\bibitem{Villani2009}
\BIBentryALTinterwordspacing
C.~Villani, \emph{The Wasserstein Distances}.\hskip 1em plus 0.5em minus
  0.4em\relax Berlin, Heidelberg: Springer Berlin Heidelberg, 2009, pp.
  93--111. [Online]. Available: \url{https://doi.org/10.1007/978-3-540-71050-9}
\BIBentrySTDinterwordspacing

\bibitem{GPSNoiseModel2022}
S.~Caldararu, H.~Zhang, I.~Mahajan, T.~Hansen, S.~Chatterjee, N.~Batagoda,
  R.~Serban, and D.~Negrut, ``Using random walks to simulate {GPS} sensing for
  applications in robotics and autonomous vehicles,''
  \url{https://sbel.wisc.edu/wp-content/uploads/sites/569/2023/03/TR-2022-02.pdf},
  2022.

\end{thebibliography}
% \end{thebibliography}

%%%%%%%%%%%%%%%%%%%%%%%%%%%%%%%%%%%%%%%%%%%%%%%%%%%%%%%%%%%%%%%%%%%%%%%%%%%%%%%%
\section*{APPENDIX}
\subsection{IMU and GPS Noise Models}
\subsubsection*{\textbf{Project Chrono}}
\label{sec:app:imu_gps_models}
The IMU model utilizes a Gaussian drift noise model for both the gyroscope and accelerometer. The gyroscope model is given by, $\vomega_{out} = \vomega + \veta_a + \vb_t,\:\: \veta_a \sim \mathcal{N}(\vmu_a, \vSigma_a)$ with $\vb_t = \vb_{t-1} + \veta_b,\:\: \veta_b \sim \mathcal{N}(\vmu_b, \vb_0\sqrt{\frac{dt}{t_b}})$.

Here $\vomega_{out} \in \mathbb{R}^3$ is the final measure of angular velocity produced by the gyroscope, $\vomega \in \mathbb{R}^3$ is the ground truth angular velocity, $\vmu_a$, $\vmu_b$, $\vSigma_a$, $\vb_0$, $dt$ and $t_b$ are user-defined parameters. We set $\vmu_a = \vmu_b = 0$, $\vSigma_a = \text{diag}(\sigma_x^2, \sigma_y^2, \sigma_z^2)$ where $\sigma_i$, $\forall i \in {x,y,z}$ is obtained from the IMU's data sheet. Additionally, we model drift by setting $\vb_0 = \{1e^{-4}, 1e^{-4}, 1e^{-4}\}$, $t_b = 0.1$ and $\vmu_b = \{0, 0, 0\}$.

The accelerometer uses the same noise model with angular velocity $\vomega$ replaced with linear acceleration $\va$. We set the drift parameters the same as the gyroscope, center the Gaussian noise around zero, and set the covariance matrix $\vSigma_a$ based on the accelerometer's data sheet.

We use two variants of the GPS sensor model in Chrono: one with a normal noise model and the other with a random walk noise model. The normal noise model is given by, $ \vp_{out} = \vp + \veta,\:\:\:\:\:\: \veta \sim \mathcal{N}(\vmu, \vSigma)$,

where $\vp_{out} \in \mathbb{R}^3$ is the final measure of position produced by the GPS, $\vp \in \mathbb{R}^3$ is the ground truth position, $\vmu = 0$, and $\vSigma = \text{diag}(\sigma_x^2, \sigma_y^2, \sigma_z^2)$ where $\sigma_i$, $\forall i \in {x,y,z}$ is obtained from the GPS's data sheet. 
The random walk noise model uses a normal distribution with a mean $\vmu$ and a covariance of $\vSigma_a = \text{diag}(\sigma_{ax}^2, \sigma_{ay}^2, \sigma_{az}^2)$ in acceleration~\cite{GPSNoiseModel2022}. To prevent the GPS from drifting too far away from the ground truth position, the mean $\vmu$ is set based on a concentration gradient that provides an acceleration in a direction forcing the GPS to return to the ground truth position. The acceleration is then integrated to obtain the velocity and position. The random walk noise model is given by, $\vmu_{t+1} = \frac{\bar{\vp}}{\bar{\vp}_{\text{max}}}$ used in $\va_{t+1} \sim \mathcal{N}(\vmu_{t+1}, \vSigma_a)$ integrated to provide $\bar{\vp}_{t+1} = \bar{\vp}_{t} + \vv_{t}dt + \frac{1}{2}\va_{t+1}dt^2$.

Here $\bar{\vp}$ is the accumulated error in the GPS position, $\bar{\vp}_{\text{max}}$ is the maximum allowable error set by the user, $\va_{t+1}$ is the acceleration of the error at time $t+1$ with a mean $\vmu_{t+1}$ and a covariance $\vSigma_a$ set by the user, $\vv_{t}$ is the velocity of the error at time $t$, and $dt$ is the time step set by the user. Finally, the accumulated error in the GPS position is updated at the GPS's data publication frequency as, $\vp_{out} = \vp + \bar{\vp}$. Where $\vp_{out}$ is the final measure of position produced by the GPS, and $\vp$ is the ground truth position. The standard deviation $\vSigma_a$ is calculated by empirically measuring the standard deviation of the second derivative of the real GPS's position using a standstill test. $\bar{\vp}_{\text{max}}$ is set to $0.06$ m based on the empirical measurements of the GPS's position noise on the same standstill test. 

\subsubsection*{\textbf{AirSim}}
\label{sec:app:airsim_imu_gps_models}
The IMU model in AirSim is the same as that in Chrono. The GPS noise model on the other hand does not add any noise to the GPS position which is given by querying the simulator. Instead, the model computes the horizontal dilution of precision (HDOP) which is a measure of the geometric quality of a GPS satellite configuration in the sky~\cite{Jwo2001GPSHDOP,Zhong2006Geometric}. The HDOP value in reality thus depends on many factors, such as the time of day, weather conditions etc. AirSim does not model the position and movements of satellite but rather employs a low pass filter to model reduction in HDOP over time which is given by, $\alpha = \exp \left(\frac{-\Delta T}{\tau}\right)$ used in $H_{t+1} = \alpha H_{t} + (1-\alpha) H_{\infty}$. Here $\alpha$ is the filter coefficient, $\Delta T$ is the time step defined by the GPS frequency, $\tau$ is the time constant of the filter, $H_{t}$ is the HDOP at time $t$, and $H_{\infty}$ is the HDOP at convergence. $H_0$, the initial HDOP is set to $100$, representing a poor GPS signal at the start. The time constant $\tau$ and the HDOP value at convergence $H_{\infty}$ and calibrated based on the real-world GPS's HDOP values. The HDOP that is obtained from the sensor model is converted into a diagonal covariance matrix $\vSigma_H$ by the GPS sensor driver in reality. Since there is no need to use a sensor driver in simulation, we implement the calculation of $\vSigma_H$ in the sensor model itself and is given by; 
\begin{align}
    \vSigma_H &= \text{diag}(\sigma_x^2, \sigma_y^2, \sigma_z^2)
    \label{eq:airsim_cov}
\end{align}
Here $\sigma_i = (0.02H_i)^2 \:\:\:\forall i \in {x,y,z}$. The constant $0.02$ is obtained directly from the GPS driver code that is used in reality.

\label{sec:app}

\end{document}